\documentclass[10pt,twocolumn,letterpaper]{article}

\PassOptionsToPackage{table}{xcolor} 

\usepackage[pagenumbers]{iccv} 

\usepackage[accsupp]{axessibility}  

\usepackage{times}
\usepackage{colortbl}
\usepackage{epsfig}
\usepackage{graphicx}
\usepackage{amsmath}
\usepackage{amssymb}
\usepackage{makecell} 

\usepackage{enumitem} 
\usepackage[linesnumbered,ruled,vlined,algo2e]{algorithm2e}
\usepackage{algorithm, algorithmic}
\usepackage{multirow}
\usepackage{threeparttable}
\usepackage{stfloats}
\usepackage{booktabs}
\usepackage{arydshln}
\usepackage{bm}
\usepackage{makecell}
\usepackage{pifont}
\usepackage{lipsum}
\usepackage{ulem}
\definecolor{iccvblue}{rgb}{0.21,0.49,0.74}
\usepackage[pagebackref,breaklinks,colorlinks,allcolors=iccvblue]{hyperref}

\usepackage{mathtools} 
\usepackage{lmodern}
\usepackage{appendix}
\usepackage{booktabs}
\usepackage{enumitem}

\usepackage[capitalize]{cleveref}
\crefname{section}{Sec.}{Secs.}
\Crefname{section}{Section}{Sections}
\Crefname{table}{Table}{Tables}
\crefname{table}{Tab.}{Tabs.}

\colorlet{colorFst}{Green!25}       
\colorlet{colorSnd}{SpringGreen!45} 
\colorlet{colorTrd}{Yellow!30}      
\colorlet{colorLow}{darkgray!30}    

\newcommand\blfootnote[1]{%
  \begingroup
  \renewcommand\thefootnote{}\footnote{#1}%
  \addtocounter{footnote}{-1}%
  \endgroup
}

\def\paperID{2695} 
\def\confName{ICCV}
\def\confYear{2025}

\title{SGAD: Semantic and Geometric-aware Descriptor for Local Feature Matching}

\author{
Xiangzeng Liu$^{1*}$, Chi Wang$^{1* \dagger}$, Guanglu Shi$^{1}$, Xiaodong Zhang$^{1}$, Qiguang Miao$^{1\dagger}$, Miao Fan$^{2}$ \\
$^{1}$Xidian University \quad $^{2}$Navinfo Europe B.V \\
{\small Project page: \href{https://mr-chiwang.github.io/SGAD/}{https://mr-chiwang.github.io/SGAD/}}
}

\begin{document}
\maketitle
\begin{abstract}
    Local feature matching remains a fundamental challenge in computer vision.
    Recent Area to Point Matching (A2PM) methods have improved matching accuracy. 
    However, existing research based on this framework relies on inefficient pixel-level comparisons and complex graph matching that limit scalability.
    In this work, we introduce the Semantic and Geometric-aware Descriptor Network (SGAD), which fundamentally rethinks area-based matching by generating highly 
    discriminative area descriptors that enable direct matching without complex graph optimization. 
    This approach significantly improves both accuracy and efficiency of area matching. 
    We further improve the performance of area matching through a novel supervision strategy 
    that decomposes the area matching task into classification and ranking subtasks. 
    Finally, we introduce the Hierarchical Containment Redundancy Filter (HCRF) to eliminate overlapping areas by analyzing containment graphs.
    SGAD demonstrates remarkable performance gains, reducing runtime by 60$\times$ (0.82s$~vs.~$60.23s) compared to MESA. 
    Extensive evaluations show consistent improvements across multiple point matchers: 
    SGAD+LoFTR reduces runtime compared to DKM, while achieving higher accuracy (0.82s$~vs.~$1.51s, 65.98$~vs.~$61.11) in outdoor pose estimation, 
    and SGAD+ROMA delivers +7.39\% AUC@5$^\circ$ in indoor pose estimation, establishing a new state-of-the-art.
\end{abstract}    

\blfootnote{$^{*}$Equal Contribution. $^{\dagger}$Corresponding Author.}

\section{Introduction}
\label{sec:intro}
Feature matching establishes precise pixel-level correspondences between images, which is fundamental to numerous computer vision tasks, 
including SFM~\cite{sfm}, SLAM~\cite{orb_slam3}, visual localization~\cite{sattler2017cvpr,localization}, and image retrieval~\cite{retrieval}. 
Nevertheless, achieving accurate pixel correspondences remains challenging due to scale variations, viewpoint changes, illumination differences, and repetitive patterns.

\begin{figure}[!t]
    \centering
    \includegraphics[width=\columnwidth]{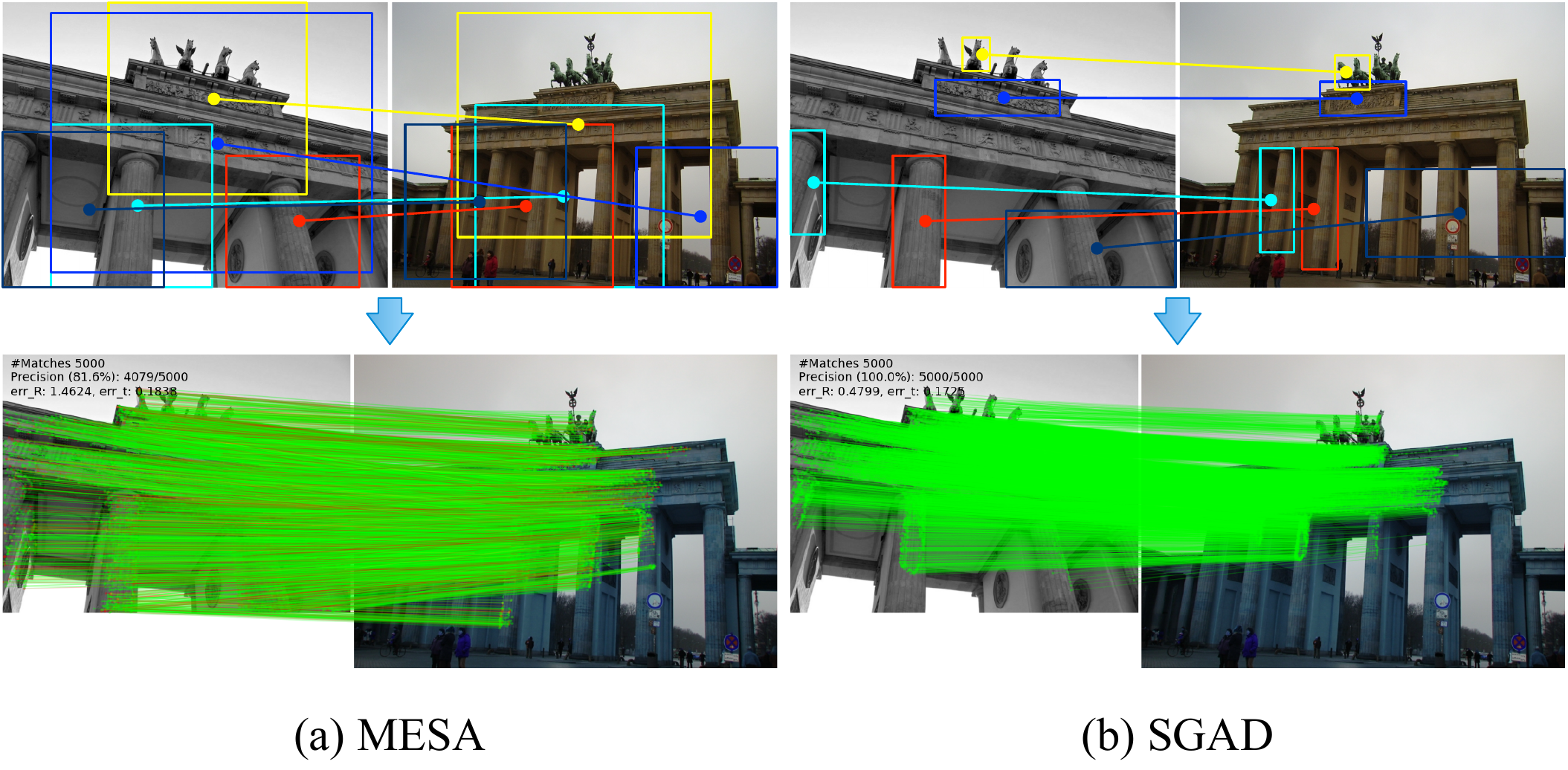}
    \caption{
    Area matching comparison between MESA and our proposed SGAD. SGAD accurately identifies highly coherent corresponding areas, establishing a stronger foundation for subsequent pixel-level matching.
    }
    \label{fig:area_matching}
\end{figure}

Most existing methods process entire image pairs through extensive pixel comparisons, increasing computational complexity and often leading to 
mismatches due to the involvement of many non-correlated pixels. 
Recent research has addressed this by identifying overlapping areas in advance and restricting matching to those areas. 
Approaches such as~\cite{adamatcher,OETR} segment overlapping areas before matching, though implicit learning introduces non-reusable computational overhead. 
TopicFM~\cite{topicfm} classifies pixels into topics, limiting matching to pixels within the same topic. 
SGAM~\cite{sgam} introduced the Area to Point Matching (A2PM) framework, improving correspondence accuracy by matching semantic areas before point matching, yet its implementation relies on explicit semantic labels, making it vulnerable to annotation inaccuracies.

MESA~\cite{mesa} addresses the dependency on semantic labels by incorporating the Segment Anything Model (SAM)~\cite{sam}. 
However, it suffers from two critical limitations: high computational cost due to inefficient pixel-level comparisons and complex graph matching, 
and reduced accuracy from non-overlapping areas introduced by its merging strategy. 
Similarly, DMESA~\cite{dmesa} enhances matching efficiency by generating a dense matching distribution, 
yet still struggles with the computational burden of graph matching and area merging limitations of MESA. 
As shown in~\cref{fig:area_matching}(a), MESA's area correspondences can suffer from content inconsistencies, which hinders subsequent refinement.

To overcome these fundamental limitations, we propose a novel paradigm: the Semantic and Geometric-aware Descriptor Network (SGAD). 
Our approach generates highly discriminative area descriptors that enable straightforward matching with a simple descriptor matcher, 
eliminating the need for complex graph optimization. This approach is significantly more efficient and scalable.
Specifically, SGAD first segments images into areas using SAM~\cite{sam} and then extracts initial semantic features for each area via DINOv2~\cite{dinov2}, providing rich contextual information about the area content.
To address the lack of spatial context in these features, a geometric positional encoding module embeds spatial relationship information, 
reducing interference from visually similar areas at different positions. 
Subsequently, alternating self-attention and cross-attention mechanisms~\cite{attention} model complex relationships between areas both within and across images. 
Additionally, our Hierarchical Containment Redundancy Filter (HCRF) constructs containment relations graphs and filters out redundant areas based on node relationships. 
For training, we introduce a novel dual-task supervision strategy that decomposes area matching into classification and ranking. This joint optimization enables the model to learn both absolute (classification) and relative (ranking) similarity patterns.
As shown in~\cref{fig:area_matching}(b), SGAD produces more content-consistent area matching results than MESA, 
providing a stronger foundation for subsequent fine-grained pixel-level matching.

The key contributions of our work are as follows:

1) A novel area matching paradigm (SGAD) that achieves efficient and accurate area matching, 
addressing the efficiency and scalability limitations of prior A2PM frameworks (\cref{subsec:area_description,subsec:descriptor_matcher}).

2) A dual task learning framework decomposing area matching into classification and ranking subtasks, jointly optimizing both absolute and 
relative similarity patterns (\cref{subsec:supervision}, validated in~\cref{tab:ablation}).

3) Consistent performance improvements across multiple point matchers on indoor and outdoor datasets. 
SGAD+LoFTR (semi-dense) outperforms DKM (dense) in both efficiency and accuracy on MegaDepth ($0.82s~vs.~1.51s$, $65.98~vs.~61.11$).
Furthermore, our method establishes a new state-of-the-art when integrated with ROMA.
\section{Related Work}
\noindent \textbf{Classic Feature Matching.}
Feature matching methods can be broadly classified into detector-based and detector-free approaches. Detector-based methods follow a three-stage pipeline: 
feature detection~\cite{sift,keyNet,superpoint}, description~\cite{d2net,r2d2}, and matching~\cite{superglue,lightglue,OANet}. 
While feature detection reduces the search space, these methods struggle with large viewpoint changes and texture-less scenes. 
In contrast, detector-free methods~\cite{loftr,topicfm,dkm,roma} enhance keypoint repeatability by directly extracting dense visual descriptors and 
generating dense matches.
However, the large-scale dense comparisons incur significant computational overhead, and often lead to a higher rate of false matches.

\noindent \textbf{Multi-stage Feature Matching.}
To reduce redundant matches, researchers have explored various strategies. 
Methods like~\cite{patch2pix, pats} adopted uniform patch division but overlooked image semantic structure.
TopicFM~\cite{topicfm} addressed this by introducing a topic model-based pixel classification method, 
though its generalizability and cross-image consistency remain limited. 
These approaches highlight the value of semantics, but its potential remains largely underutilized for efficient and accurate matching.

\noindent \textbf{Area to Point Matching Methods.}
The A2PM framework, introduced by SGAM~\cite{sgam}, enhances performance by restricting point matching to semantically matched areas. 
However, SGAM heavily relies on predefined semantic labels, making it sensitive to annotation quality.
MESA~\cite{mesa} mitigated this dependency by leveraging the SAM~\cite{sam}, but introduced new challenges: 
dense activity map computation and complex graph-matching optimization created substantial computational overhead, while 
area merging often introduced non-overlapping areas into the initial segmentation, degrading matching quality. 
Although DMESA improved efficiency through dense matching distributions, it still inherits the fundamental limitations of graph-matching and area merging from MESA.

Our method fundamentally differs from previous approaches by rethinking area-based matching through a dedicated descriptor network. 
Prior work often relies on complex matching algorithms to compensate for inadequate area representations. 
In contrast, SGAD focuses on generating highly discriminative descriptors from the outset. This enables straightforward matching without complex graph optimization. 
This paradigm shift significantly improves both matching accuracy and computational efficiency, 
addressing the core limitations of existing A2PM approaches.
\section{Method}
\label{sec:method}

\begin{figure*}[htbp]
    \centering
    \includegraphics[width=\textwidth]{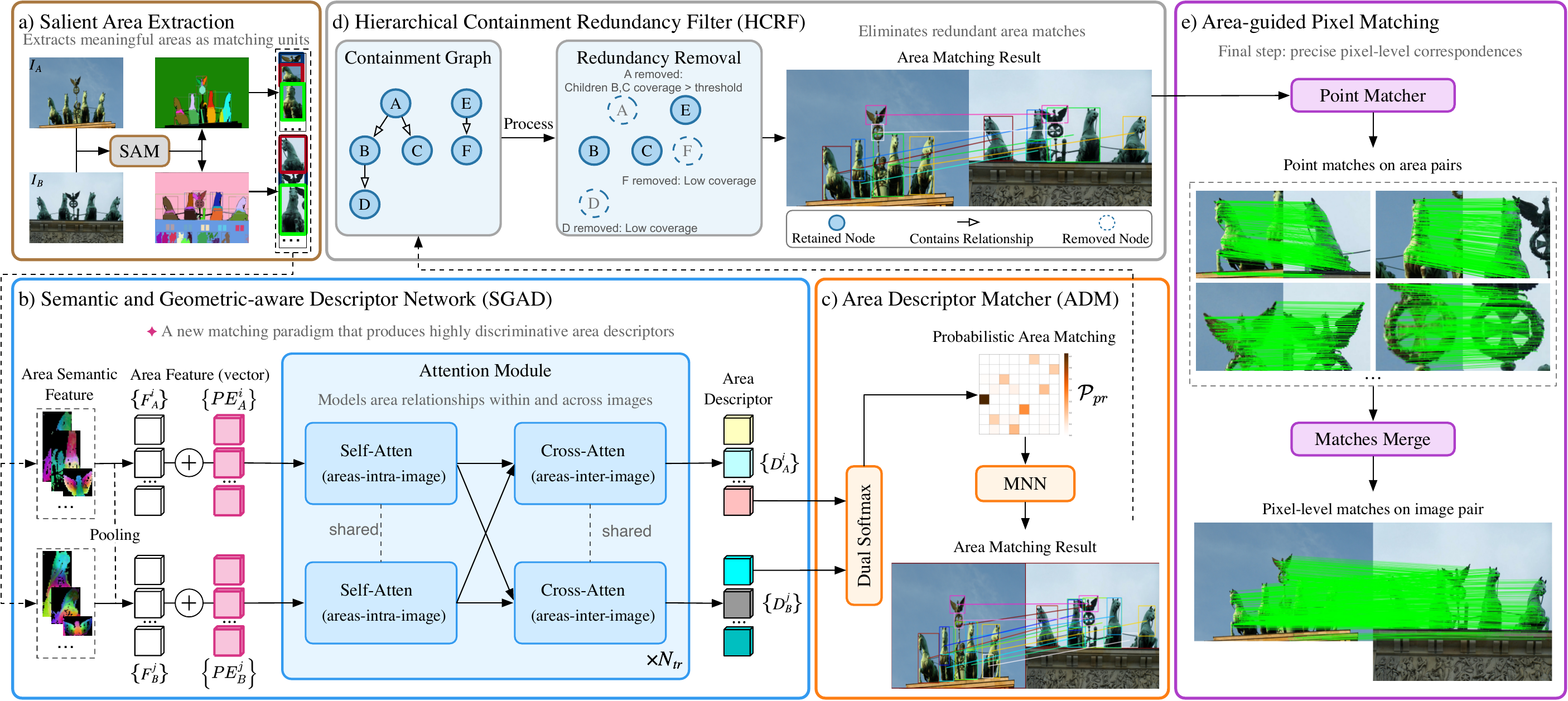}
    \caption{Overview of the SGAD method. SGAD consists of five components: 
    (a) Salient areas are extracted from image pairs using SAM.
    (b) Semantic features are extracted by DINOv2 and average-pooled into vectors \(\{F_A^i\}\) and \(\{F_B^j\}\).
    Spatial relationships are encoded through geometric positional embeddings \(\{PE_A^i\}\) and \(\{PE_B^j\}\). 
    Alternating self-attention and cross-attention model complex relationships between areas both within and across images,
    producing final descriptors \(\{D_A^i\}\) and \(\{D_B^j\}\).
    (c) Dual-softmax estimation is applied to compute the area matching probability matrix \(\mathcal{P}_{pr}\), 
    with the Mutual Nearest Neighbor (MNN) algorithm selecting matched pairs \(\mathcal{M}\). 
    (d) Hierarchical Containment Redundancy Filter (HCRF) removes nested areas from \(\mathcal{M}\), retaining non-redundant matched pairs.
    (e) Local feature matching is performed within selected matched area pairs, with final pixel-level matching obtained by aggregating all matched points across areas.
    }
    \label{fig:sgad}
\end{figure*}

\cref{fig:sgad} illustrates the architecture of our proposed Semantic and Geometric-aware Descriptor Network (SGAD). 
SGAD first establishes matches between salient areas and then guides precise pixel-level matching within these areas. 
The framework consists of five key components: (a) salient area extraction, (b) area description through our semantic and geometric-aware descriptor network, 
(c) area matching with a simple descriptor matcher, (d) hierarchical containment redundancy filter, and (e) guided point matching within matched areas.

\subsection{Salient Area Extraction and Description} \label{subsec:area_description}
\noindent \textbf{Area Feature Extraction.}
We leverage SAM to generate instance masks for \(I_A\) and \(I_B\), identifying the minimum 
bounding rectangles of these masks as extracted areas, forming sets \(R_A = \{R_A^i \mid i = 1, 2, \dots, m\}\) and \(R_B = \{R_B^j \mid j = 1, 2, \dots, n\}\) 
as illustrated in~\cref{fig:sgad}(a). The indices \(i\) and \(j\) used later will follow these same ranges.
Next, we use a frozen DINOv2 encoder to extract semantically rich features for each area, 
which are then average-pooled to obtain feature vectors \(\{F_A^i\}\) and \(\{F_B^j\}\), 
significantly reducing input length for subsequent attention operations.

\noindent \textbf{Geometric Positional Encoding.}
Although \(\{F_A^i\}\) and \(\{F_B^j\}\) capture strong semantic features, they lack global spatial information, as each feature is extracted independently. 
Such global spatial context is crucial for accurate area matching. 
To address this, we designed a geometric positional encoding module based on relative geometric attributes between areas.
Given the two sets of areas, \(\{R_A^i\}\) and \(\{R_B^j\}\), with center coordinates \((x_A^i, y_A^i)\) and \((x_B^j, y_B^j)\), 
we compute the geometric spatial attributes of each area relative to all other areas within the same image, including the Euclidean distances 
and relative angles between them.

The Euclidean distance is given by:
\begin{equation}
\left\{
\begin{alignedat}{2}
    &d_{A}^{il} &&= \sqrt{(x_A^i - x_A^l)^2 + (y_A^i - y_A^l)^2} \\
    &d_{B}^{jk} &&= \sqrt{(x_B^j - x_B^k)^2 + (y_B^j - y_B^k)^2},
\end{alignedat}
\right.
\end{equation}    
where \(l\) and \(k\) are the intra-image area indices for Image A and Image B, respectively.

The relative angle is computed using the arctangent:
\begin{equation}
\left\{
\begin{alignedat}{2}
    &\theta_{A}^{il} &&= \operatorname{atan2}(y_A^i - y_A^l, x_A^i - x_A^l) \\
    &\theta_{B}^{jk} &&= \operatorname{atan2}(y_B^j - y_B^k, x_B^j - x_B^k).
\end{alignedat}
\right.
\end{equation}

The positional embedding for each area is then generated by an MLP. Its input is the average of the geometric vectors, \([d^{il}, \sin(\theta^{il}), \cos(\theta^{il})]\), computed pairwise with all other areas in the same image:   
\begin{equation}
    \begin{cases}
        PE_A^i = \operatorname{MLP}\left( \frac{1}{m-1} \sum_{l \neq i}^{m} [d_{A}^{il}, \sin(\theta_{A}^{il}), \cos(\theta_{A}^{il})] \right) \\[0.5em]
        PE_B^j = \operatorname{MLP}\left( \frac{1}{n-1} \sum_{k \neq j}^{n} [d_{B}^{jk}, \sin(\theta_{B}^{jk}), \cos(\theta_{B}^{jk})] \right).
    \end{cases}
\end{equation}

Then, these positional embeddings are added to the area features:
\begin{equation}
\left\{
\begin{alignedat}{2}
    &\hat{F}_A^i &&= F_A^i + PE_A^i \\
    &\hat{F}_B^j &&= F_B^j + PE_B^j.
\end{alignedat}
\right.
\end{equation}
    
This approach enhances spatial representation by integrating relative positions and geometric relationships.

\noindent \textbf{Attention Module.}
To model complex relationships between areas both within and across images, we adopt an attention mechanism enabling both intra-image and inter-image information interaction.
For the area feature sequences 
\(\{\hat{F}_A^i\} = [\hat{F}_A^1, \hat{F}_A^2, \dots, \hat{F}_A^m]\) and \(\{\hat{F}_B^j\} = [\hat{F}_B^1, \hat{F}_B^2, \dots, \hat{F}_B^n]\), 
we first apply self-attention separately within \(\{\hat{F}_A^i\}\) and \(\{\hat{F}_B^j\}\), 
followed by cross-attention between them. By stacking \(N_{tr}\) layers of 
alternating self-attention and cross-attention, we obtain final area descriptors \(\{D_A^i\}\) and \(\{D_B^j\}\) with improved expressiveness 
and discriminative power of each area feature.


\subsection{Area Descriptor Matcher} \label{subsec:descriptor_matcher}
\textbf{Matching Probabilities Estimation.} 
We compute the score matrix \( \mathcal{S} \) between the area descriptors:
\begin{align}
    \mathcal{S}(i, j) = \frac{1}{\tau} \cdot \langle D_A^i, D_B^j \rangle,
\end{align}
where \( \tau \) is a temperature parameter adjusting score sharpness.

Dual-softmax~\cite{NCN,disk} estimates the probability matrix for matching area descriptors. The matching probability \( \mathcal{P}_{pr} \) for a pair \((i, j)\) is:
\begin{align}
    \mathcal{P}_{pr}(i, j) = \text{softmax}(\mathcal{S}(i, \cdot))_j \cdot \text{softmax}(\mathcal{S}(\cdot, j))_i.
\end{align}

\noindent \textbf{Match Selection.}
Based on confidence matrix \( \mathcal{P}_{pr} \), we select matches with confidence values above threshold \( \lambda_{pr} \). 
To enforce one-to-one matching and exclude potential outliers, we apply the Mutual Nearest Neighbor (MNN) algorithm to find optimal assignment. 
The resulting set of area match predictions is:
\begin{align}
    \mathcal{M} = \{ (i, j) \mid (i, j) \in \text{MNN}(\mathcal{P}_{pr}),\; \mathcal{P}_{pr}(i, j) \geq \lambda_{pr} \}.
\end{align}
\subsection{Hierarchical Containment Redundancy Filter} \label{subsec:Filter}
\begin{figure}[!t]
    \centering
    \includegraphics[width=\columnwidth]{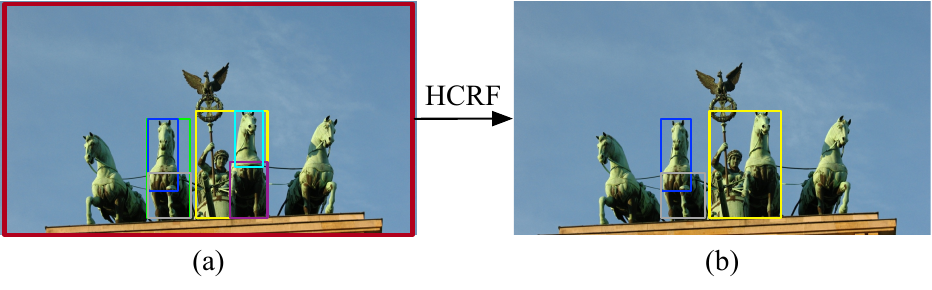}
    \caption{Visualization of the HCRF process. 
    (a) Matching results directly obtained through MNN, with overlapping areas present. 
    (b) Results after filtering through HCRF.
    HCRF effectively resolves the problem of overlapping areas, preserving the optimal matching results.}

    \label{fig:hcrf}
\end{figure}

Descriptor matching yields matched area pairs as shown in~\cref{fig:sgad}(c). 
While these pairs accurately identify matching relationships even for multi-level nested areas, 
extensively overlapping areas introduce significant redundant computations in subsequent point matching (\cref{fig:hcrf}(a)). 
To address this, we propose a Hierarchical Containment Redundancy Filter (HCRF) to eliminate this redundancy effectively.

We first construct an area containment relationship graph (\cref{fig:sgad}(d)), representing soft containment controlled by 
threshold $\delta_{\text{contain}}$. The containment relationship is defined as:
\begin{align}
    R_p \text{ contains } R_c \text{ iff } \frac{|R_p \cap R_c|}{|R_c|} \geq \delta_{\text{contain}},
\end{align}
where $|R_p \cap R_c|$ is the overlapping area and $|R_c|$ is the size of the contained area.

We traverse the containment graph using Depth-First-Search (DFS)~\cite{dfs} and determine which nodes to retain based on coverage ratio and threshold $\delta_{\text{cover}}$. For parent node $R_p$ and its child nodes $C(p)$:
\begin{equation}
    \begin{aligned}
        &\text{If } \frac{|\cup_{c \in C(p)} R_c|}{|R_p|} < \delta_{\text{cover}}, \text{ keep } R_p; \\
        &\text{otherwise, keep } \{R_c | c \in C(p)\},
    \end{aligned}
\end{equation}
where $\cup_{c \in C(p)} R_c$ is the union of all child areas and $|R_p|$ is the parent area size.

We apply this filter on one image (e.g., Image A) and use the resulting indices to prune the match set $\mathcal{M}$.
The resulting filtered indices identify the most representative area pairs while eliminating redundancy, as shown in~\cref{fig:hcrf}(b), 
significantly improving the efficiency of subsequent point matching.

We deliberately perform filtering after area matching for two key reasons. First, SGAD is highly efficient and completes area matching quickly even without pre-filtering. Second, filtering prematurely could aggressively prune areas in some cases, reducing the number of matched pairs and thus harming the final point matching performance.

\subsection{Supervision.} \label{subsec:supervision}
To optimize area matching performance, we design a novel dual-task framework that decomposes the process into classification and ranking subtasks. 
The classification subtask determines whether area pairs meet basic matching criteria, establishing a foundation for viable matches. 
The ranking subtask refines this process by selecting the best match from multiple candidates, capturing relative similarity patterns. 
Through joint optimization with \( \mathcal{L} = \mathcal{L}_{cls} + \mathcal{L}_{rank} \), our model effectively captures both absolute and relative similarity patterns, ensuring robust matching.

\noindent \textbf{Area Matching Label.}
Following SuperGlue~\cite{superglue}, we compute precise pixel-level correspondences using camera poses and depth maps. 
The ground-truth matching score $\mathcal{P}_{\text{gt}}(i, j)$ is the Intersection over Union (IoU)~\cite{IoU} between area $R_B^j$ and the projection of area $R_A^i$ into Image B, denoted as $\text{proj}(R_A^i)$, using the ground-truth camera poses:
\begin{align}
    \mathcal{P}_{\text{gt}}(i, j) = \frac{|\text{proj}(R_A^i) \cap R_B^j|}{|\text{proj}(R_A^i) \cup R_B^j|}.
\end{align}

Area pairs with \(\mathcal{P}_{\text{gt}}(i, j)\) above threshold \( \lambda_{gt} \) are considered similar. 
To create a binary classification ground-truth matrix \( \mathcal{P}_{\text{gt}}^{cls} \), we binarize the confidence matrix \( \mathcal{P}_{\text{gt}} \) as:
\begin{align}
    \mathcal{P}_{\text{gt}}^{cls}(i, j) = \begin{cases} 
    1, & \text{if } \mathcal{P}_{\text{gt}}(i, j) > \lambda_{gt} \\
    0, & \text{otherwise}.
    \end{cases}
\end{align}

\noindent \textbf{Classification Supervision.}
For the classification task, we employ focal loss~\cite{focalloss} with the predicted confidence matrix \( \mathcal{P}_{pr} \) obtained through dual softmax. 
The classification loss, comprising positive and negative sample components, is defined as:
\begin{align}
    \mathcal{L}_{cls} = -\alpha \cdot (1 - p_t)^\gamma \cdot \log(p_t),
\end{align}
where:
\begin{align}
p_t = \begin{cases} 
      \mathcal{P}_{pr}(i, j) & \text{if } \mathcal{P}_{\text{gt}}^{cls}(i, j) = 1 \\
      1 - \mathcal{P}_{pr}(i, j) & \text{if } \mathcal{P}_{\text{gt}}^{cls}(i, j) = 0,
      \end{cases}
\end{align}
\( \alpha \) controls the weight of positive and negative samples, and \( \gamma \) controls the weight of hard and easy samples.

\noindent \textbf{Ranking Supervision.}
For the ranking task, we adopt ListMLE loss~\cite{listmle}. 
For each source image area, we extract matching score sequences between this area and all target image areas from both \( \mathcal{P}_{pr} \) and \( \mathcal{P}_{\text{gt}} \), 
yielding predicted score sequence \( T_{pr} = [T_{pr_1}, T_{pr_2}, \dots, T_{pr_n}] \) and ground-truth score sequence \( T_{gt} = [T_{gt_1}, T_{gt_2}, \dots, T_{gt_n}] \). 
Our objective is for \( T_{pr} \) to match the ranking order of \( T_{gt} \), ensuring matches with higher ground-truth scores receive higher predicted confidence. 
The ranking loss is defined as:
\begin{align}
    \mathcal{L}_{rank} = - \sum_{k=1}^{n} \left( T_{pr, \pi^*_k} - \log \sum_{l=k}^{n} \exp(T_{pr, \pi^*_l}) \right),
\end{align}
where \( \pi^* \) represents the permutation of indices that sorts the ground-truth scores \( T_{gt} \) in descending order, and \( T_{pr, \pi^*_k} \) is the predicted score from the sequence \( T_{pr} \) at the \(k\)-th position of this permutation.

\section{Experiments}
\label{sec:experiments}


\subsection{Implementation Details}  
We train separate models for indoor and outdoor environments on the ScanNet~\cite{scannet} and MegaDepth~\cite{megadepth} datasets, respectively, with the same training-testing division as in~\cite{loftr}. 
The model was trained end-to-end using the AdamW~\cite{adamw} optimizer with an initial learning rate of \(1 \times 10^{-4}\) and a batch size of $64$. 
Training time is approximately $2$ days on a single RTX A6000 GPU. 
For feature extraction, we use DINOv2 Large~\cite{dinov2} as the backbone. 
\( N_{tr}\) is set to $4$, and the ground-truth matching threshold \( \lambda_{gt} \) is set to $0.2$.

\subsection{Area Matching}
For the matching task, we assess performance on the ScanNet1500~\cite{scannet} and MegaDepth1500~\cite{megadepth} benchmark datasets.

\begin{figure}[!t]
    \centering
    \includegraphics[width=\columnwidth]{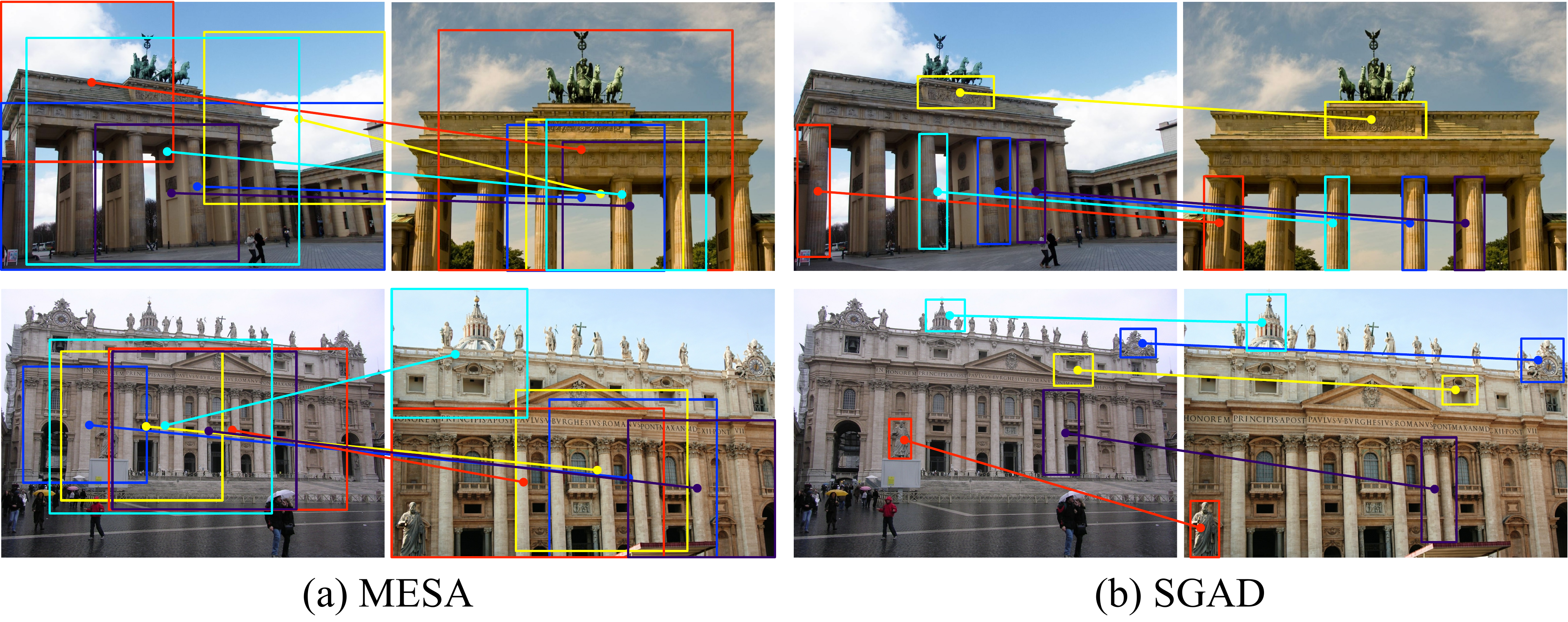}
    \caption{Comparison of area matching results between the proposed method SGAD and MESA. 
    To provide a clear comparison, we randomly selected 5 area pairs for each method.
    }
    \label{fig:area_matching_exp}
\end{figure}

\noindent \textbf{Experimental setup.}
On both indoor and outdoor datasets, we use the Area Under the Curve (AUC) metric to assess performance in correctly identifying matched and non-matched pairs, 
evaluated across multiple overlap rate thresholds (AUC@$0.2$, AUC@$0.3$, AUC@$0.4$, AUC@$0.5$).
MESA and DMESA do not provide similarity matrices for all regions involved in the matching process, 
it is not possible to calculate this metric for these methods. 
Therefore, in addition to quantitative results, we provide qualitative comparisons in~\cref{fig:area_matching_exp}.

\noindent \textbf{Results.}
As shown in~\cref{tab:Area_AUC}, SGAD achieves exceptional area matching accuracy, with AUC scores consistently above $95\%$ across all thresholds 
for both indoor and outdoor environments.~\cref{fig:area_matching_exp} demonstrates that our method achieves more accurate area matching results compared to MESA, 
particularly in challenging scenarios with large viewpoint changes. 
This robust matching capability directly contributes to improved results in downstream tasks, as evidenced by subsequent tables (\cref{tab:HPE_AUC,tab:Pose_AUC_SN,tab:Pose_AUC_MD}).

\begin{table}[t]
    \raggedright
    \resizebox{\linewidth}{!}{ 
    \begin{tabular}{lcccc}
    \toprule
    \multirow{1}{*}{ScanNet} 
                            & AUC@0.2$\uparrow$ & AUC@0.3$\uparrow$ & AUC@0.4$\uparrow$  & AUC@0.5$\uparrow$ \\ \midrule
    SGAD                     & 95.46 & 96.18 & 96.78 & 97.28 \\ \midrule
    \multirow{1}{*}{MegaDepth} 
                            & AUC@0.2$\uparrow$ & AUC@0.3$\uparrow$ & AUC@0.4$\uparrow$  & AUC@0.5$\uparrow$ \\ \midrule
    SGAD                     & 97.39 & 97.81 & 98.02 & 98.18 \\ \midrule
    \end{tabular}
    }
    \caption{Area matching results of SGAD on ScanNet1500 and MegaDepth1500. Measured in AUC (higher is better).}
    \label{tab:Area_AUC}
\end{table}

\begin{table}[t]
    \centering
    \resizebox{\linewidth}{!}{ 
    \begin{tabular}{lccc}
    \toprule
    Method & AUC@3px$\uparrow$ & AUC@5px$\uparrow$ & AUC@10px$\uparrow$ \\ 
    \midrule
    SP\cite{superpoint}+SG\cite{superglue}~\tiny{CVPR'19}   & 53.9 & 68.3 & 81.7 \\ 
    LoFTR~\cite{loftr}~\tiny{CVPR'21}                      & 65.9 & 75.6 & 84.6 \\ 
    3DG-STFM~\cite{3dgstfm}~\tiny{ECCV'22}                  & 64.7 & 73.1 & 81.0 \\ 
    AspanFormer~\cite{aspanformer}~\tiny{ECCV'22}           & 67.4 & 76.9 & 85.6 \\ 
    TopicFM~\cite{topicfm}~\tiny{AAAI'23}                   & 67.3 & 77.0 & 85.7 \\ 
    PDCNet+~\cite{pdcnetplus}~\tiny{PAMI'23}                  & 67.7 & 77.6 & 86.3 \\ 
    \midrule
    DKM~\cite{dkm}~\tiny{CVPR'23}                           & \underline{71.3} & \underline{80.6} & \underline{88.5} \\ 
    SGAD+DKM                                                & \textbf{72.1} & \textbf{81.3} & \textbf{89.2} \\ 
    \bottomrule
    \end{tabular}
    }
    \caption{Homography estimation results on HPatches. Measured in AUC (higher is better).}
    \label{tab:HPE_AUC}
\end{table}
\subsection{Homography Estimation}
We evaluate SGAD on the HPatches dataset for homography estimation, a standard benchmark containing diverse viewpoint changes and illumination variations.
Following~\cite{superglue} and~\cite{loftr}, we report AUC at three thresholds ($3$, $5$, and $10$ pixels), with all images preprocessed by resizing their shorter side to $480$. 
We compare against recent approaches~\cite{superpoint,superglue,loftr,3dgstfm,aspanformer,topicfm,pdcnetplus,dkm}.

\noindent \textbf{Results.}
As shown in~\cref{tab:HPE_AUC}, SGAD significantly improves upon the strong DKM baseline, increasing AUC@3px from $71.3\%$ to $72.1\%$. 
Unlike MESA and DMESA, which fail to establish area matches in certain scenes and are therefore excluded from comparison, 
SGAD maintains robust performance across all test cases. 
This demonstrates the superior ability of our method to handle challenging matching scenarios where previous area-based approaches typically struggle.
\begin{table*}[t]
    \centering
    \resizebox{\linewidth}{!}{
    \begin{tabular}{ccclllllllll}
    \toprule
    \multicolumn{3}{l}{\multirow{2}{*}{Pose estimation AUC}} & \multicolumn{9}{c}{ScanNet1500 benchmark} \\ \cmidrule(l){4-12} 
    \multicolumn{3}{c}{}                        & \multicolumn{3}{c}{$1296 \times 968$} & \multicolumn{3}{c}{$880 \times 640$} & \multicolumn{3}{c}{$640 \times 480$} \\ \cmidrule(l){4-6} \cmidrule(l){7-9} \cmidrule(l){10-12} 
    \multicolumn{3}{c}{}                        & AUC@5$^\circ\uparrow$ & AUC@10$^\circ\uparrow$ & AUC@20$^\circ\uparrow$ 
                                               & AUC@5$^\circ\uparrow$ & AUC@10$^\circ\uparrow$ & AUC@20$^\circ\uparrow$
                                               & AUC@5$^\circ\uparrow$ & AUC@10$^\circ\uparrow$ & AUC@20$^\circ\uparrow$ \\ 
    \midrule
    \multicolumn{3}{l}{TopicFM~\cite{topicfm}~\tiny{AAAI'23}}                  & 19.14 & 36.55 & 52.68 & 19.23 & 36.48 & 52.49 & 19.45 & 36.57 & 52.75 \\
    \multicolumn{3}{l}{TopicFM+~\cite{topicfmplus}~\tiny{TIP'24}}                & 20.26 & 37.83 & 54.06 & 19.73 & 37.19 & 53.64 & 20.00 & 37.79 & 53.98 \\
    \midrule
    \multicolumn{3}{l}{SP\cite{superpoint}+SG\cite{superglue}~\tiny{CVPR'19}}  & 22.62 & 42.89 & 61.44 & 23.37 & 43.68 & 62.76 & 21.73 & 41.64 & 60.41 \\
    \rowcolor[rgb]{.9,.95,.98}\multicolumn{3}{l}{SGAD+SPSG}                   & 25.74$_{+13.79\%}$ & 45.95$_{+7.13\%}$ & 63.77$_{+3.79\%}$ & 25.17$_{+7.70\%}$ & 45.46$_{+4.08\%}$ & 63.48$_{+1.15\%}$ & 24.31$_{+11.87\%}$ & 44.64$_{+7.20\%}$ & 62.92$_{+4.15\%}$ \\
    \midrule

    \multicolumn{3}{l}{LoFTR~\cite{loftr}~\tiny{CVPR'21}}                      & 8.91 & 18.31 & 29.56 & 20.01 & 39.50 & 57.70 & 28.44 & 50.43 & 68.80 \\
    \rowcolor[rgb]{.9,.95,.98}\multicolumn{3}{l}{SGAD+LoFTR}                   & 29.69$_{+233.22\%}$ & 51.50$_{+181.27\%}$ & 69.58$_{+135.39\%}$ & 28.79$_{+43.88\%}$ & 50.68$_{+28.30\%}$ & 68.92$_{+19.45\%}$ & 28.69$_{+0.88\%}$ & 51.18$_{+1.49\%}$ & 69.40$_{+0.87\%}$ \\
    \midrule
    \multicolumn{3}{l}{DKM~\cite{dkm}~\tiny{CVPR'23}}                         & 24.16 & 44.03 & 61.34 & 29.30 & 51.02 & 68.51 & 30.17 & 51.80 & 69.52 \\
    \multicolumn{3}{l}{MESA~\cite{mesa}+DKM~\tiny{CVPR'24}}                                               & {30.07}$_{+24.46\%}$ & {51.57}$_{+17.12\%}$ & {68.98}$_{+12.46\%}$ & {30.76}$_{+4.98\%}$ & {52.59}$_{+3.08\%}$ & {69.74}$_{+1.80\%}$ & 30.85$_{+2.25\%}$ & {52.57}$_{+1.49\%}$ & {69.97}$_{+0.65\%}$ \\
    \multicolumn{3}{l}{DMESA~\cite{dmesa}+DKM~\tiny{Arxiv'24}}                                               & 28.89$_{+19.58\%}$ & 49.34$_{+12.06\%}$ & 66.28$_{+8.05\%}$ & 30.61$_{+4.47\%}$ & 51.95$_{+1.82\%}$ & 69.14$_{+0.92\%}$ & {30.90}$_{+2.42\%}$ & 52.31$_{+0.98\%}$ & 69.86$_{+0.49\%}$ \\
    \rowcolor[rgb]{.9,.95,.98}\multicolumn{3}{l}{SGAD+DKM}                   & \uline{31.63}$_{+30.92\%}$ & \uline{52.98}$_{+20.33\%}$ & \uline{69.93}$_{+14.00\%}$ & \uline{31.85}$_{+8.70\%}$ & \uline{53.11}$_{+4.10\%}$ & \uline{70.12}$_{+2.35\%}$ & \uline{31.65}$_{+4.91\%}$ & \uline{53.12}$_{+2.55\%}$ & \uline{70.69}$_{+1.68\%}$ \\
    \rowcolor[rgb]{.9,.95,.98}\multicolumn{3}{l}{SGAD$^{\dagger}$+DKM}       & 31.51$_{+30.42\%}$ & 52.87$_{+20.08\%}$ & 69.87$_{+13.91\%}$ & 31.76$_{+8.40\%}$ & 53.02$_{+3.92\%}$ & 70.05$_{+2.25\%}$ & 31.49$_{+4.38\%}$ & 52.92$_{+2.16\%}$ & 70.55$_{+1.48\%}$ \\
    \midrule
    \multicolumn{3}{l}{ROMA~\cite{roma}~\tiny{CVPR'24}}                         & 31.51 & 53.44 & 71.10 & 31.99 & 53.90 & 71.39 & 31.80 & 53.92 & 71.29 \\
    \rowcolor[rgb]{.9,.95,.98}\multicolumn{3}{l}{SGAD+ROMA}                   & \textbf{33.84}$_{+7.39\%}$ & \textbf{55.37}$_{+3.61\%}$ & \textbf{72.23}$_{+1.59\%}$ & \textbf{33.61}$_{+5.06\%}$ & \textbf{55.19}$_{+2.39\%}$ & \textbf{72.11}$_{+1.01\%}$ & \textbf{33.49}$_{+5.31\%}$ & \textbf{55.14}$_{+2.26\%}$ & \textbf{72.16}$_{+1.22\%}$ \\
    
    \bottomrule
    \end{tabular}
    }
    \vspace{-0.5em}
    \caption{Relative pose estimation results (\%) on ScanNet1500. Measured in AUC (higher is better). $^{\dagger}$ denotes model trained on the MegaDepth dataset.}
    \label{tab:Pose_AUC_SN}
\end{table*}
\begin{table*}[t]
    \centering
    \resizebox{\linewidth}{!}{
    \begin{tabular}{ccclllllllll}
    \toprule
    \multicolumn{3}{l}{\multirow{2}{*}{Pose estimation AUC}} & \multicolumn{9}{c}{MegaDepth1500 benchmark} \\ \cmidrule(l){4-12} 
    \multicolumn{3}{c}{}                        & \multicolumn{3}{c}{$1200 \times 1200$} & \multicolumn{3}{c}{$832 \times 832$} & \multicolumn{3}{c}{$640 \times 640$} \\ \cmidrule(l){4-6} \cmidrule(l){7-9} \cmidrule(l){10-12} 
    \multicolumn{3}{c}{}                        & AUC@5$^\circ\uparrow$ & AUC@10$^\circ\uparrow$ & AUC@20$^\circ\uparrow$ 
                                               & AUC@5$^\circ\uparrow$ & AUC@10$^\circ\uparrow$ & AUC@20$^\circ\uparrow$
                                               & AUC@5$^\circ\uparrow$ & AUC@10$^\circ\uparrow$ & AUC@20$^\circ\uparrow$ \\ \midrule
    \multicolumn{3}{l}{SP\cite{superpoint}+SG\cite{superglue}~\tiny{CVPR'19}}  & 56.83 & 71.9 & 83.03 & 53.32 & 68.75 & 80.66 & 47.28 & 63.57 & 76.50 \\
    \multicolumn{3}{l}{TopicFM~\cite{topicfm}~\tiny{AAAI'23}}                  & 52.68 & 69.44 & 81.42 & 49.36 & 67.28 & 80.01 & 46.53 & 64.15 & 77.73 \\
    \multicolumn{3}{l}{TopicFM+~\cite{topicfmplus}~\tiny{TIP'24}}                & 56.52 & 71.93 & 82.87 & 55.03 & 70.18 & 81.49 & 49.53 & 65.31 & 77.49 \\
    \midrule
    \multicolumn{3}{l}{LoFTR~\cite{loftr}~\tiny{CVPR'21}}                      & 62.37 & 76.34 & 85.96 & 60.64 & 74.82 & 84.83 & 56.42 & 71.80 & 82.65 \\
    \rowcolor[rgb]{.9,.95,.98}\multicolumn{3}{l}{SGAD+LoFTR}                   & 65.98$_{+5.79\%}$ & \uline{78.77}$_{+3.18\%}$ & \uline{87.13}$_{+1.36\%}$ & 65.16$_{+7.45\%}$ & 77.94$_{+4.17\%}$ & 86.62$_{+2.11\%}$ & 63.94$_{+13.33\%}$ & 76.90$_{+7.10\%}$ & 85.78$_{+3.79\%}$ \\
    \midrule

    \multicolumn{3}{l}{DKM~\cite{dkm}~\tiny{CVPR'23}}                         & 61.11 & 74.63 & 84.02 & 62.42 & 75.88 & 85.11 & 63.26 & 76.13 & 84.97 \\
    \multicolumn{3}{l}{MESA~\cite{mesa}+DKM~\tiny{CVPR'24}}                                               & 62.31$_{+1.96\%}$ & 76.11$_{+1.98\%}$ & {85.56}$_{+1.83\%}$ & 62.68$_{+0.42\%}$ & 75.96$_{+0.11\%}$ & 85.35$_{+0.28\%}$ & 63.02$_{-0.38\%}$ & 76.31$_{+0.24\%}$ & 85.60$_{+0.74\%}$ \\
    \multicolumn{3}{l}{DMESA~\cite{dmesa}+DKM~\tiny{Arxiv'24}}                                               & {63.52}$_{+3.94\%}$ & {76.29}$_{+2.22\%}$ & {85.31}$_{+1.54\%}$ & {64.02}$_{+2.56\%}$ & {76.69}$_{+1.07\%}$ & {85.54}$_{+0.51\%}$ & {65.24}$_{+3.13\%}$ & {77.98}$_{+2.43\%}$ & {86.55}$_{+1.86\%}$ \\
    \rowcolor[rgb]{.9,.95,.98}\multicolumn{3}{l}{SGAD+DKM}                   & \uline{66.40}$_{+8.66\%}$ & 78.38$_{+5.02\%}$ & 86.51$_{+2.96\%}$ & \uline{66.49}$_{+6.52\%}$ & \uline{78.80}$_{+3.85\%}$ & \uline{87.23}$_{+2.49\%}$ & \uline{66.75}$_{+5.52\%}$ & \uline{78.80}$_{+3.51\%}$ & \uline{87.13}$_{+2.54\%}$ \\
    \rowcolor[rgb]{.9,.95,.98}\multicolumn{3}{l}{SGAD$^{\dagger}$+DKM}         & 66.04$_{+8.08\%}$ & 78.07$_{+4.61\%}$ & 86.50$_{+2.95\%}$ & 65.97$_{+5.69\%}$ & 78.35$_{+3.26\%}$ & 86.88$_{+2.08\%}$ & 66.73$_{+5.49\%}$ & 78.39$_{+2.97\%}$ & 86.67$_{+2.00\%}$ \\
    \midrule
    \multicolumn{3}{l}{ROMA~\cite{roma}~\tiny{CVPR'24}}                         & 65.68 & 78.15 & 86.68 & 65.91 & 78.41 & 86.95 & 65.29 & 78.01 & 86.68 \\
    \rowcolor[rgb]{.9,.95,.98}\multicolumn{3}{l}{SGAD+ROMA}                   & \textbf{67.85}$_{+3.30\%}$ & \textbf{79.87}$_{+2.20\%}$ & \textbf{88.02}$_{+1.55\%}$ & \textbf{68.34}$_{+3.69\%}$ & \textbf{80.27}$_{+2.37\%}$ & \textbf{88.34}$_{+1.57\%}$ & \textbf{67.94}$_{+4.06\%}$ & \textbf{80.09}$_{+2.67\%}$ & \textbf{88.40}$_{+1.98\%}$ \\
    
    \bottomrule
    \end{tabular}
    }
    \vspace{-0.5em}
    \caption{Relative pose estimation results (\%) on MegaDepth1500. Measured in AUC (higher is better). $^{\dagger}$ denotes model trained on the ScanNet dataset.}
    \label{tab:Pose_AUC_MD}
\end{table*}
\begin{figure*}[!t]
    \centering
    \includegraphics[width=\textwidth]{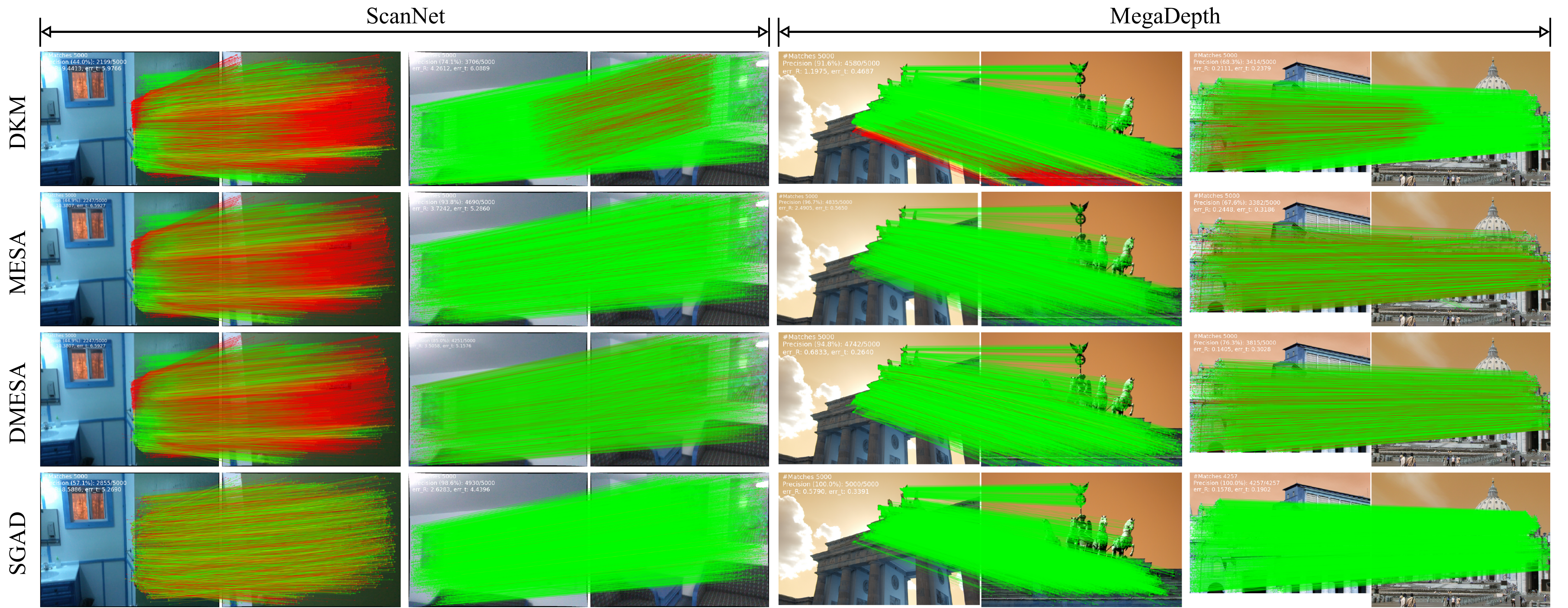}
    \caption{Qualitative comparison. We compare SGAD against MESA~\cite{mesa} and DMESA~\cite{dmesa}, with all three methods using DKM~\cite{dkm} as the downstream point matcher. SGAD achieves more correct matches and lower rotation and translation errors on scenes with large viewpoint changes (MegaDepth) and untextured scenes (ScanNet).
    }

    \label{fig:pose_all}
\end{figure*}

\subsection{Relative Pose Estimation}\label{subsec:pose}
\noindent \textbf{Datasets.}
We assess SGAD pose estimation accuracy on ScanNet1500, a challenging indoor dataset with untextured surfaces and viewpoint variations, 
and on MegaDepth1500, a benchmark of 1500 outdoor image pairs.

\noindent \textbf{Experimental setup.}
We conduct extensive experiments across various image resolutions.
For ScanNet, we resize full images to $1296 \times 968$ (original resolution), $880 \times 640$, and $640 \times 480$. 
For MegaDepth, we test at resolutions of $1200 \times 1200$, $832 \times 832$, and $640 \times 640$ for full images.
For all experiments, cropped areas are consistently resized to $640 \times 480$ (ScanNet) or $640 \times 640$ (MegaDepth).
Following \cite{dmesa}, we implement MAGSAC++\cite{magsac} for outlier rejection. 
We evaluate the performance improvement of SGAD with various point matchers and compare it against recent approaches~\cite{superpoint, superglue, loftr, topicfm, topicfmplus, dkm, roma, mesa, dmesa}.
We evaluate pose estimation using AUC at thresholds $(5^\circ, 10^\circ, 20^\circ)$\cite{superglue}.
To assess cross-domain generalization capability, we conduct two additional experiments: 
testing SGAD on ScanNet using a model trained on MegaDepth (outdoor→indoor transfer), and testing on MegaDepth using a model trained on ScanNet (indoor→outdoor transfer). 
In both cases, we maintain point matching models trained on their respective test domains.

\noindent \textbf{Results.}
\cref{tab:Pose_AUC_SN,tab:Pose_AUC_MD} demonstrate that SGAD substantially enhances the performance of all baseline methods across all resolutions for both indoor and outdoor scenes.
Notably, in indoor scenes (ScanNet), at $1296 \times 968$ resolution, SGAD achieves a $+233.22\%$ improvement in AUC@5$^\circ$ for LoFTR, 
effectively addressing the matching difficulties in high-resolution indoor scenes.
Moreover, compared to other methods based on the A2PM framework such as MESA and DMESA, SGAD achieves superior improvement, 
e.g., $+30.92\%$ in AUC@5$^\circ$ for DKM at $1296 \times 968$ resolution, outperforming both MESA ($24.46\%$) and DMESA ($19.58\%$).
Additionally, in cross-domain generalization experiments for both indoor and outdoor scenes, SGAD achieves performance comparable to models trained on the test domain.
Furthermore, as shown in~\cref{fig:pose_all}, SGAD achieves more correct matches and lower rotation and translation errors on untextured scenes and with large viewpoint changes.
These results demonstrate that SGAD effectively enhances the performance of various point matchers while exhibiting strong generalization ability and robustness.
Significantly, when combined with the SOTA method ROMA, SGAD provides significant improvement across all resolutions in both indoor and outdoor scenes, 
with SGAD+ROMA establishing a new \textit{state-of-the-art}.

\subsection{Runtime}\label{subsec:runtime}
We conducted comprehensive timing experiments on both indoor and outdoor datasets to evaluate computational efficiency, with an RTX A6000 GPU.
For all methods tested, images from the MegaDepth dataset were resized to $1200 \times 1200$ pixels, while images from the ScanNet dataset 
retained their original dimensions ($1296 \times 968$ pixels). The area size was consistently resized to $480 \times 480$ pixels.
The SAM extraction step is offline, so we do not include it in the timing.

\noindent \textbf{Results.}
As shown in~\cref{tab:runtime}, SGAD demonstrated significant computational efficiency improvements in both indoor and outdoor scenes compared to previous A2PM methods.
The processing times on MegaDepth of $0.82s$ for SGAD+LoFTR versus $60.23s$ for MESA+LoFTR and $1.84s$ for DMESA+LoFTR. 
Furthermore, our analysis in~\cref{tab:Pose_AUC_MD,tab:runtime} reveals that SGAD+LoFTR (semi-dense) not only achieved 
lower processing times than DKM (dense) ($0.82s~vs.~1.51s$) but also consistently outperformed it across multiple resolution levels, 
delivering superior accuracy with AUC@5$^\circ$ values of $65.98~vs.~61.11$ at $1200 \times 1200$ resolution 
and $65.16~vs.~62.42$ at $832 \times 832$ resolution, 
thereby demonstrating that our approach provides both computational efficiency and performance advantages in diverse scenarios. 

\subsection{Ablation Study}
\label{sec:ablation}

We perform ablation studies on ScanNet to validate our design choices (\cref{tab:ablation}).
1) The DINOv2-only baseline achieves 79.74 AUC@0.2.
2) \& 3) Sequentially adding the attention mechanism and positional encoding (PE) progressively boosts performance to 92.52 and finally 95.46, demonstrating that all architectural components are essential and the gains do not merely stem from the powerful backbone.
4) \& 5) The study also highlights our supervision strategy's effectiveness: removing the ranking loss (\(\mathcal{L}_{rank}\)) degrades performance, while replacing our dual-task loss with a standard Triplet loss causes a significant drop to 91.81 AUC. This confirms the superiority of our proposed framework.

\begin{figure}[!t]
    \centering
    \includegraphics[width=\columnwidth]{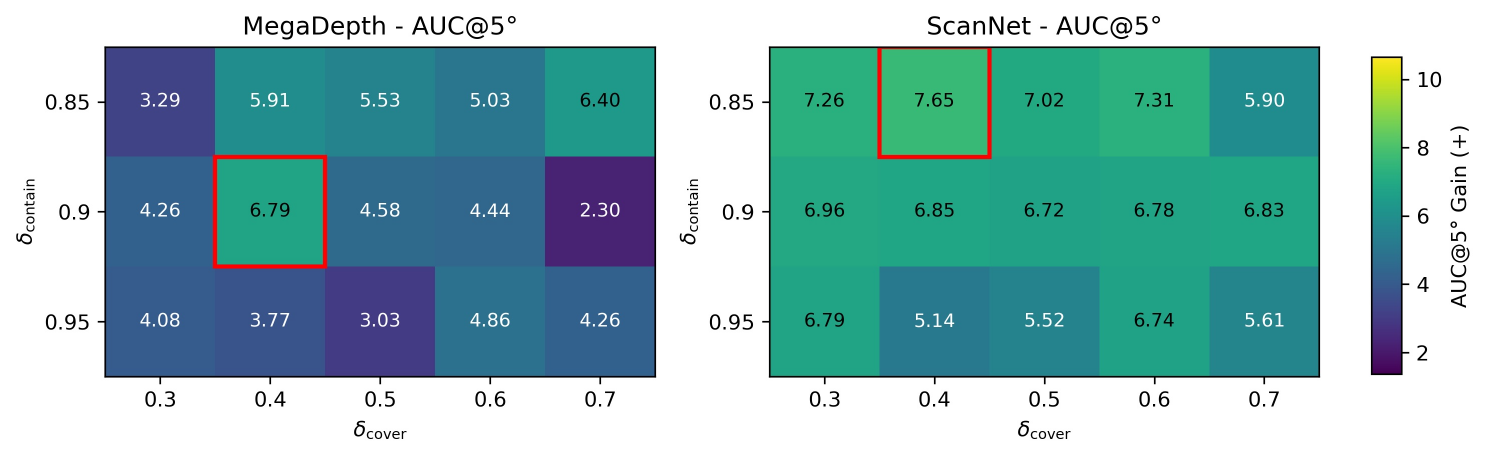}
    \caption{
    Sensitivity analysis of the HCRF module. The heatmaps plot the absolute AUC@5° gain of SGAD over the LoFTR baseline for pose estimation. Results are shown for MegaDepth (640x640) and ScanNet (880x640).    }
    \label{fig:ablation_hcrf}
\end{figure}

We analyze the sensitivity of our HCRF parameters on a randomly sampled subset of 300 image pairs. As shown in~\cref{fig:ablation_hcrf}, we achieve the best AUC@5$^\circ$ gains of +6.79 at (0.4, 0.9) on MegaDepth and +7.65 at (0.4, 0.85) on ScanNet. Parameters within the ranges of Coverage=[0.4-0.5] and Overlap=[0.85-0.9] perform well across both datasets.

\begin{table}[t]
    \centering
    \resizebox{\linewidth}{!}{
    \begin{tabular}{lccccc} 
    \toprule
                & LoFTR & DKM & MESA+LoFTR & DMESA+LoFTR & SGAD+LoFTR\\ \midrule
    MegaDepth & 0.38 & 1.51 & 60.23  & 1.84 & 0.82 \\
    ScanNet & 0.28 & 0.72 & 33.44 & 1.38 & 0.67 \\
    \bottomrule
    \end{tabular}
    }
    \caption{Runtime Comparison on an RTX A6000 GPU (seconds).
    SGAD includes Descriptor Network, Descriptor Matching, and HCRF.
    }
    \label{tab:runtime}
  \end{table}

\begin{table}[t]
    \centering
    \resizebox{\linewidth}{!}{
    \begin{tabular}{p{0.2cm}cccccccc} 
    \toprule
                & DINOv2 & Attention & PE & \(\mathcal{L}_{cls}\) & \(\mathcal{L}_{rank}\) & \(\mathcal{L}_{Triplet}\) & AUC @0.2$\uparrow$ & AUC @0.3$\uparrow$ \\ \midrule
    \textbf{1)} & \ding{51} & \ding{55}  & \ding{55}            & \ding{51} & \ding{51} & \ding{55}    & 79.74 & 82.62 \\
    \textbf{2)} & \ding{51} & \ding{51}  & \ding{55}            & \ding{51} & \ding{51} & \ding{55}    & 92.52 & 94.23 \\
    \textbf{3)} & \ding{51} & \ding{51}  & \ding{51}            & \ding{51} & \ding{51} & \ding{55}    & \textbf{95.46} & \textbf{96.18} \\
    \textbf{4)} & \ding{51} & \ding{51}  & \ding{51}            & \ding{51} & \ding{55} & \ding{55}    & 94.82 & 95.61 \\
    \textbf{5)} & \ding{51} & \ding{51}  & \ding{51}            & \ding{55} & \ding{55} & \ding{51}    & 91.81 & 93.55 \\
    \bottomrule
    \end{tabular}
    }
    \caption{Ablation study. Comparison of different variants of SGAD trained and evaluated on the ScanNet1500.}
    \label{tab:ablation}
\end{table}

\section{Conclusion}
\label{sec:conclusion}
We introduced SGAD, a new paradigm for Area to Point Matching (A2PM). Our approach redefines the task by learning to generate discriminative, holistic descriptors for image areas, enabling direct and efficient matching. This circumvents the bottlenecks of traditional methods, which rely on costly dense feature comparisons or complex graph-based optimization. 
This synergistic design proves both robust and efficient. Extensive experiments show that SGAD not only achieves state-of-the-art accuracy but also significantly improves computational efficiency. Its robust performance gains across multiple point matchers validate its potential as a new foundational component for feature matching pipelines.

\noindent \textbf{Limitations and Future Work.}
While SGAD demonstrates strong performance, certain limitations remain.
First, pooling area features into compact vectors limits their expressive power in challenging cases that combine high visual similarity with extreme geometric transformations, leading to matching failures (see supplementary material). 
Second, despite HCRF filtering, partial overlaps can persist among the matched areas, causing redundant downstream computations. 
Finally, the independent training of the area matching network precludes its joint, end-to-end optimization with the point matcher. 
Addressing these limitations will be the focus of our future work.

\section{Acknowledgments}
\label{sec:acknowledgments}
This work was supported by the Natural Science Basic Research Program of Shaanxi (2024JC-YBMS-467). We thank the reviewers for their valuable feedback, as well as Guanglu Shi for his help with our experiments. We also gratefully acknowledge the authors of LoFTR~\cite{loftr} and MESA~\cite{mesa} for their open-source codebases which were instrumental to this work.

{
    \small
    \bibliographystyle{ieeenat_fullname}
    \bibliography{main}
}


\clearpage
\setcounter{page}{1}
\maketitlesupplementary
\appendices

\section{Time Complexity Analysis}
This section investigates the computational complexity of MESA~\cite{mesa} and SGAD. 
Given source and target images containing \( M \) and \( N \) areas, respectively, with each area resized to \( H \times W \) pixels, our analysis focuses on the similarity computation stage. 

\subsection{MESA}
MESA formulates area matching as a graph matching problem, where each area in the source and target images is represented as a graph node.  
In the first stage, activity maps are computed by applying self-attention and cross-attention on feature maps that have been downsampled to \( \frac{1}{8} \) of the original resolution. 
The similarity of each paired area \( (R_A^i, R_B^j) \) is then determined as the product of their activity map expectations.  
The complexity is:
\begin{align}
\mathcal{O}(L \times M \times N \times ((H' \times W')^2 \times D)),
\end{align}
where \( H' = \frac{H}{8} \), \( W' = \frac{W}{8} \), \( L \) is the number of attention layers, and \( D \) is the feature dimension.

To reduce this complexity, MESA employs Area Bayesian Network optimization to filter areas, reducing the number of areas to \( M' \) and \( N' \). The complexity becomes:
\begin{align} \label{eq:sim_mesa}
\mathcal{O}(L \times M' \times N' \times ((H' \times W')^2 \times D)),
\end{align}
where \( M' < M \) and \( N' < N \).  

The subsequent graph matching has a computational complexity of: 
\begin{align}
\mathcal{O}(M^2 + N^2).
\end{align}

\subsection{SGAD}
SGAD adopts a distinct strategy. For each area, it generates a compact descriptor by first pooling the corresponding DINOv2 features into a single vector. This initial descriptor is subsequently refined through \( N_{tr} \) layers of self-attention and cross-attention.
The computational complexity is:
\begin{align} \label{eq:sim_sgad}
\mathcal{O}(N_{tr} \times M \times N \times D),
\end{align}
where \( D \) represents the feature dimension, which is independent of the image resolution.  

After generating the confidence matrix \(\mathcal{P}_{pr}\), SGAD uses the MNN algorithm to compute the matching results. The complexity is:
\begin{align}
\mathcal{O}(M \times N).
\end{align}

\begin{table}[t]
    \centering
    \resizebox{\linewidth}{!}{
    \begin{tabular}{lll}
    \toprule
    Method & Step & Complexity \\ 
    \midrule
    \multirow{3}{*}{MESA~\cite{mesa}} 
    & Similarity Calculation & \(\mathcal{O}(L \times M' \times N' \times (H' \times W')^2 \times D)\) \\
    & Graph Matching & \(\mathcal{O}(M^2 + N^2)\) \\
    & Main Bottleneck & \(\mathcal{O}(L \times M' \times N' \times (H' \times W')^2 \times D)\) \\
    \midrule
    \multirow{3}{*}{SGAD} 
    & Similarity Calculation & \(\mathcal{O}(N_{tr} \times M \times N \times D)\) \\
    & Descriptor Matching & \(\mathcal{O}(M \times N)\) \\
    & Main Bottleneck & \(\mathcal{O}(N_{tr} \times M \times N \times D)\) \\
    \bottomrule
    \end{tabular}
    }
    \caption{Complexity analysis of SGAD and MESA~\cite{mesa}.}
    \label{tab:complexity_analysis}
\end{table}

\subsection{Comparative Analysis}
As shown in~\cref{tab:complexity_analysis}, the primary computational complexity of both SGAD and MESA lies in the similarity calculation module. 
MESA computes similarity node by node, requiring a complexity as described in~\cref{eq:sim_mesa}, 
where the pixel-based activity map computation limits parallelism, and the cost remains tied to high-resolution image features. 
In contrast, SGAD's approach is independent of image resolution, which not only lowers the theoretical complexity but also enables more efficient hardware implementation, as discussed next.

In the matching stage, SGAD utilizes the Mutual Nearest Neighbor (MNN) algorithm, which operates directly on the dense confidence matrix \(\mathcal{P}_{pr}\), 
The MNN algorithm benefits from efficient GPU parallelization due to its simplicity and the dense matrix structure, making it highly scalable for large-scale tasks.
By contrast, the graph matching in MESA relies on iterative optimizations over sparsely connected graphs, which inherently limits its scalability and efficiency, especially when dealing with a large number of areas.

To validate the theoretical analysis, we measured the runtime performance of MESA and SGAD under varying numbers of areas (AreaNum).  
As shown in~\cref{tab:Time_comparison}, the empirical results closely align with the theoretical predictions.  
When AreaNum increases from 11.18 to 30.89, the runtime of MESA increases by more than 6 times (from 49.92s to 311.63s).  
In contrast, when AreaNum increases from 15.13 to 36.44, SGAD exhibits only a minor runtime increase of approximately 36\% (from 0.25s to 0.34s), 
demonstrating its scalability and computational efficiency in handling large-scale tasks.  

Ultimately, the significant performance gap observed in~\cref{tab:Time_comparison} stems from these fundamental architectural differences. 
The reliance of MESA on pixel-based activity maps and node-by-node matching leads to significant bottlenecks, particularly in high-density scenarios.  
By contrast, the area descriptor approach and dense matrix computations employed by SGAD drastically reduce complexity.  
Its ability to fully exploit GPU parallelization underscores its efficiency and suitability for large-scale tasks.

\begin{table}[t]
    \centering
    \resizebox{\linewidth}{!}{%
    \begin{tabular}{lccc}
    \toprule
    Method & Time(s)$\downarrow$ & AreaNum & AreaMatchesNum \\ 
    \midrule
    MESA~\cite{mesa} & 49.92   & 11.18 & 7.42 \\
    SGAD             & \textbf{0.25} & 15.13 & 10.44 \\ 
    \midrule
    MESA~\cite{mesa} & 311.63 & 30.89 & 19.91 \\
    SGAD             & \textbf{0.34} & 36.44 & 21.54 \\
    \bottomrule
    \end{tabular}
    }
    \caption{Runtime comparison for the area matching stage on the MegaDepth1500 benchmark. The results highlight the superior efficiency of SGAD compared to MESA~\cite{mesa}.}
    \label{tab:Time_comparison}
\end{table}

\begin{table}[t]
    \centering
    \resizebox{\linewidth}{!}{
    \begin{tabular}{ccclllllllll}
    \toprule
    \multicolumn{3}{l}{\multirow{2}{*}{Pose AUC}} & \multicolumn{9}{c}{MegaDepth1500 benchmark(image size 1200x1200)} \\ \cmidrule(l){4-12} 
    \multicolumn{3}{c}{}                        & \multicolumn{3}{c}{832x832 (area size)} & \multicolumn{3}{c}{640x640 (area size)} & \multicolumn{3}{c}{480x480 (area size)} \\ \cmidrule(l){4-6} \cmidrule(l){7-9} \cmidrule(l){10-12} 
    \multicolumn{3}{c}{}                        & @5$^\circ\uparrow$ & @10$^\circ\uparrow$ & @20$^\circ\uparrow$ 
                                               & @5$^\circ\uparrow$ & @10$^\circ\uparrow$ & @20$^\circ\uparrow$
                                               & @5$^\circ\uparrow$ & @10$^\circ\uparrow$ & @20$^\circ\uparrow$ \\ 
    \midrule
    \multicolumn{3}{l}{LoFTR}                      & 61.49 & 75.47 & 85.27 & 61.49 & 75.47 & 85.27 & 61.49 & 75.47 & 85.27 \\
    \multicolumn{3}{l}{SGAD+LoFTR}                   & \textbf{66.24} & \textbf{78.40} & \textbf{86.75} & \textbf{65.10} & \textbf{77.90} & \textbf{86.44} & \textbf{65.12} & \textbf{77.86} & \textbf{86.56} \\
  
    \midrule
    \multicolumn{3}{l}{DKM}                         & 61.11 & 74.63 & 84.02 & 61.11 & 74.63 & 84.02 & 61.11 & 74.63 & 84.02 \\
    \multicolumn{3}{l}{SGAD+DKM}                   & \textbf{66.40} & \textbf{78.38} & \textbf{86.51} & \textbf{65.97} & \textbf{78.02} & \textbf{86.38} & \textbf{65.91} & \textbf{78.27} & \textbf{86.52} \\
    \midrule
    
    \multicolumn{3}{l}{ROMA}                         & 65.68 & 78.15 & 86.68 & 65.68 & 78.15 & 86.68 & 65.68 & 78.15 & 86.68 \\
    \multicolumn{3}{l}{SGAD+ROMA}                   & \textbf{68.43} & \textbf{80.35} & \textbf{88.26} & \textbf{68.12} & \textbf{80.24} & \textbf{88.14} & \textbf{67.17} & \textbf{79.07} & \textbf{87.32} \\
    \bottomrule
    \end{tabular}
    }
    \vspace{-0.5em}
    \caption{Relative pose estimation results (\%) on MegaDepth1500. Measured in AUC (higher is better).
    The baseline methods (LoFTR, DKM, ROMA) were evaluated on the full-resolution images, and their results are presented across all columns for direct comparison.
    }
    \label{tab:Pose_AUC_MD_Addition}
  \end{table}

\section{Effect of Area Size on Different Point Matchers} 
In this section, we analyze the impact of different area sizes on the performance of various point matchers on the MegaDepth1500 benchmark.
Image size resized to \(1200 \times 1200\).
For the area sizes, we tested \(832 \times 832\), \(640 \times 640\), and \(480 \times 480\).
As shown in~\cref{tab:Pose_AUC_MD_Addition}, SGAD significantly improves the performance of LoFTR, DKM, and Roma across multiple area sizes.
This further demonstrates the effectiveness of SGAD in improving the performance of different point matchers.

\section{Failure Cases}
\begin{figure}[!t]
    \centering
    \includegraphics[width=\columnwidth]{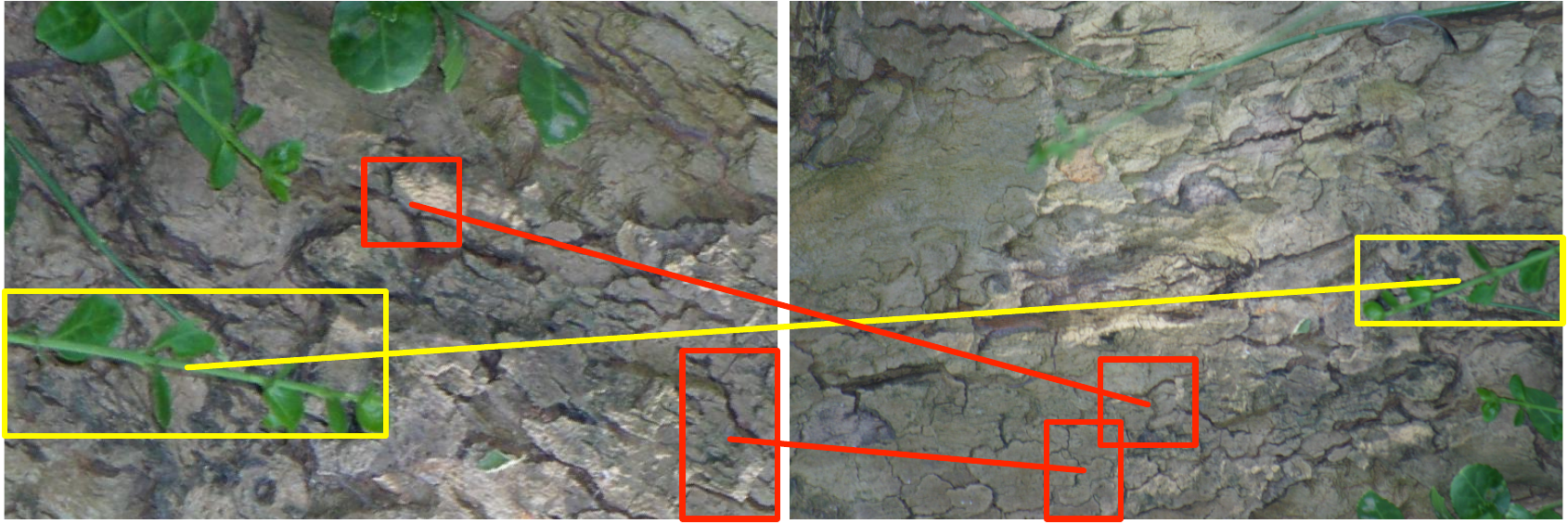}
    \caption{Failure cases of SGAD, demonstrating its vulnerability in challenging cases that combine high visual similarity and extreme geometric transformations. Mismatched areas are shown in red, correct matches in yellow.}
    \label{fig:failure_cases}
\end{figure}

\noindent In~\cref{fig:failure_cases}, we illustrate SGAD's primary failure mode, which arises in challenging cases that combine high visual similarity with extreme geometric transformations. The red color indicates mismatched areas, while the yellow color highlights correctly matched ones.

This failure stems from a detrimental synergy between our model's architecture and its training data. Our architecture, prioritizing global context via DINOv2 and pooling, inherently sacrifices the fine-grained local features required to distinguish between visually similar areas. This architectural limitation becomes particularly critical because the training data (MegaDepth and ScanNet) lacks sufficient examples of extreme geometric transformations. Consequently, the model is not explicitly trained to be robust against such distortions. When confronted with them, it must rely more heavily on the very local details that the architecture has already discarded, leading to inevitable matching failures.

\section{Additional Qualitative Results}
\subsection{Area Matching}
\begin{figure*}[htbp]
    \centering
    \includegraphics[width=\textwidth]{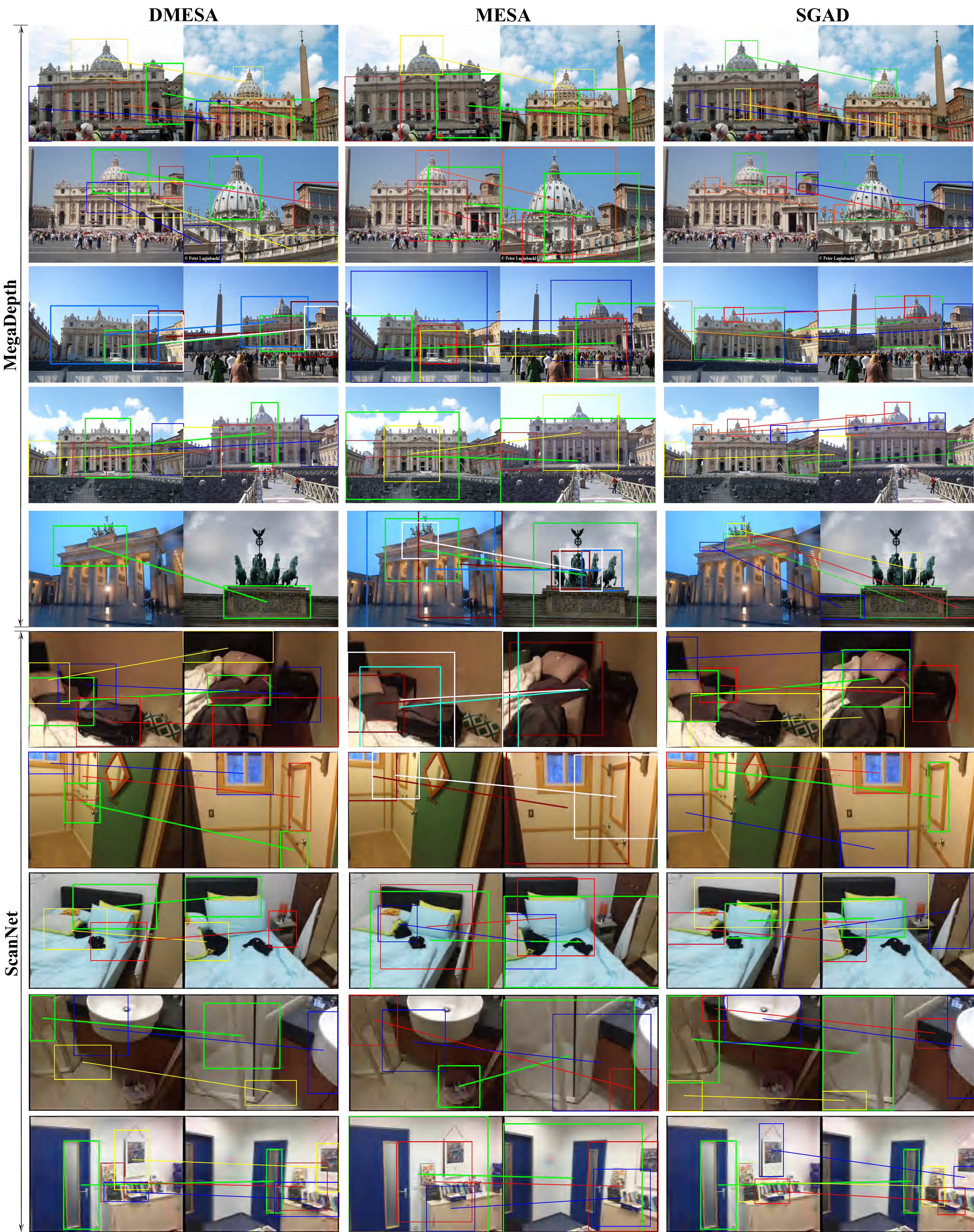}
    \caption{
    Qualitative comparison of area matching results on the MegaDepth and ScanNet datasets. 
    Our method (SGAD) is compared against MESA~\cite{mesa} and DMESA~\cite{dmesa}.
    The visualizations show that SGAD consistently identifies more semantically coherent area pairs, providing a better foundation for subsequent fine-grained matching.
    }
    \label{fig:suppl_area_matching}
\end{figure*}

This section provides a qualitative comparison of area matching between SGAD, MESA~\cite{mesa}, and DMESA~\cite{dmesa}. As shown in~\cref{fig:suppl_area_matching}, SGAD consistently finds more content-consistent area matches across the MegaDepth and ScanNet datasets. This improved consistency establishes a stronger foundation for subsequent pixel-level matching.

\subsection{Relative Pose Estimation}
\begin{figure*}[htbp]
    \centering
    \includegraphics[width=\textwidth]{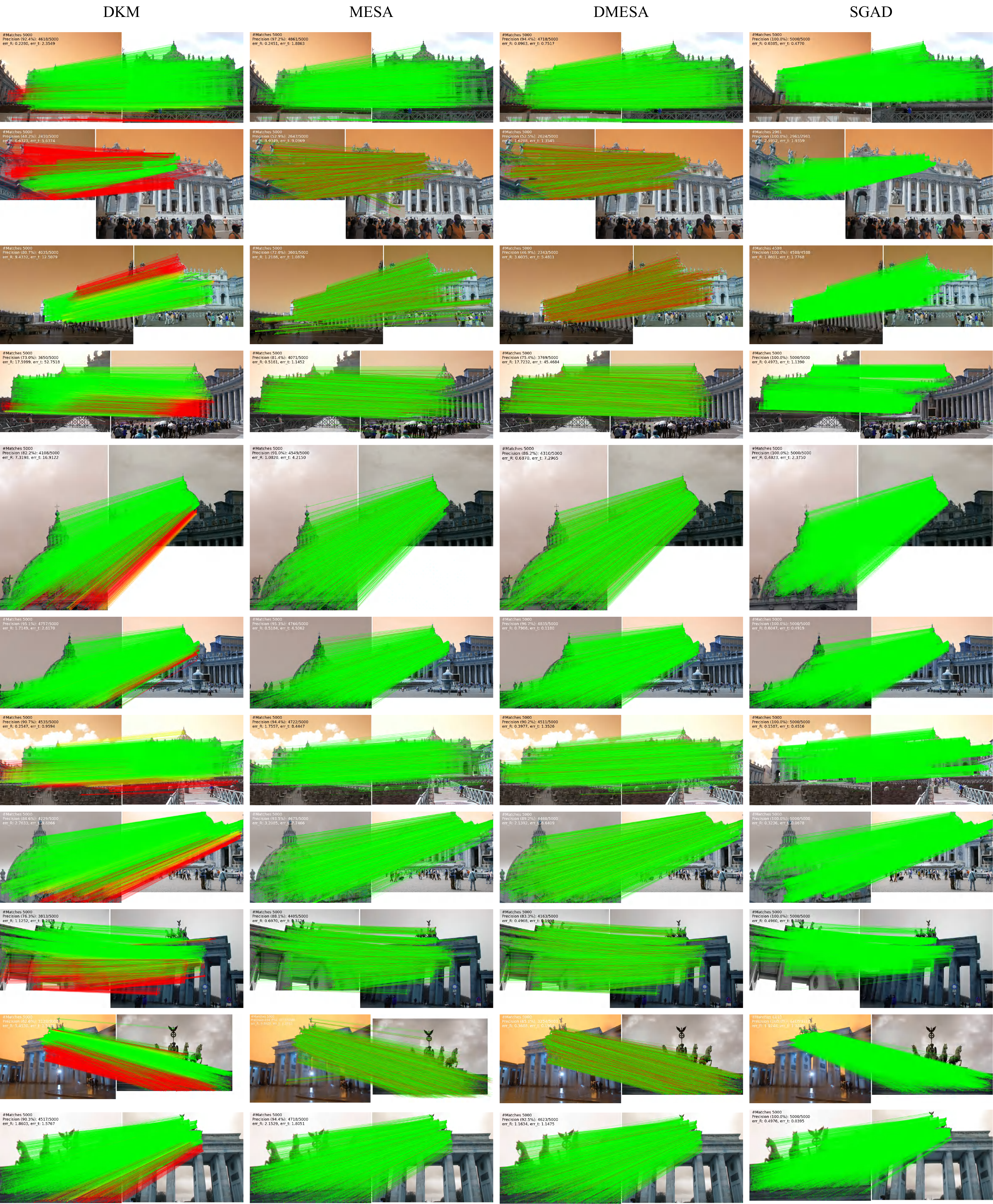}
    \caption{
    Qualitative comparison of matching methods on the MegaDepth dataset. 
    Our method (SGAD) is compared against MESA~\cite{mesa}, DMESA~\cite{dmesa}, and DKM~\cite{dkm}.
    To ensure a fair comparison of the upstream area matchers, SGAD, MESA, and DMESA all use DKM as the downstream point matcher, while DKM is also evaluated as a standalone baseline.
    Matches with an epipolar error greater than \(1 \times 10^{-4}\) are highlighted in red. 
    The results show SGAD leads to more correct final matches and lower pose errors.
    }
    \label{fig:suppl_md}
\end{figure*}

\begin{figure*}[htbp]
    \centering
    \includegraphics[width=\textwidth]{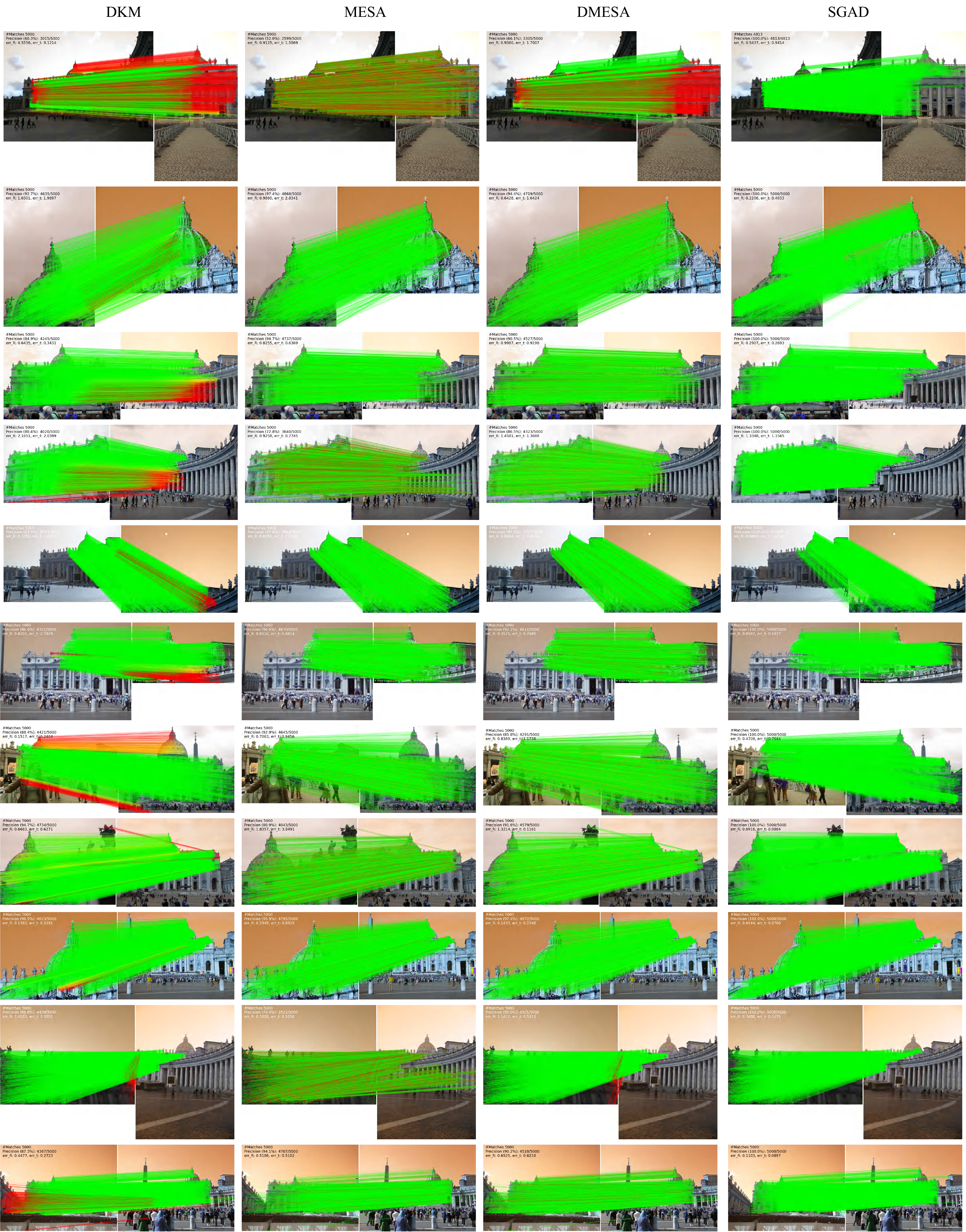}
    \caption{Qualitative comparison of matching methods on the MegaDepth dataset (continued).}
    \label{fig:suppl_md_continued}
\end{figure*}

\begin{figure*}[htbp]
    \centering
    \includegraphics[width=\textwidth]{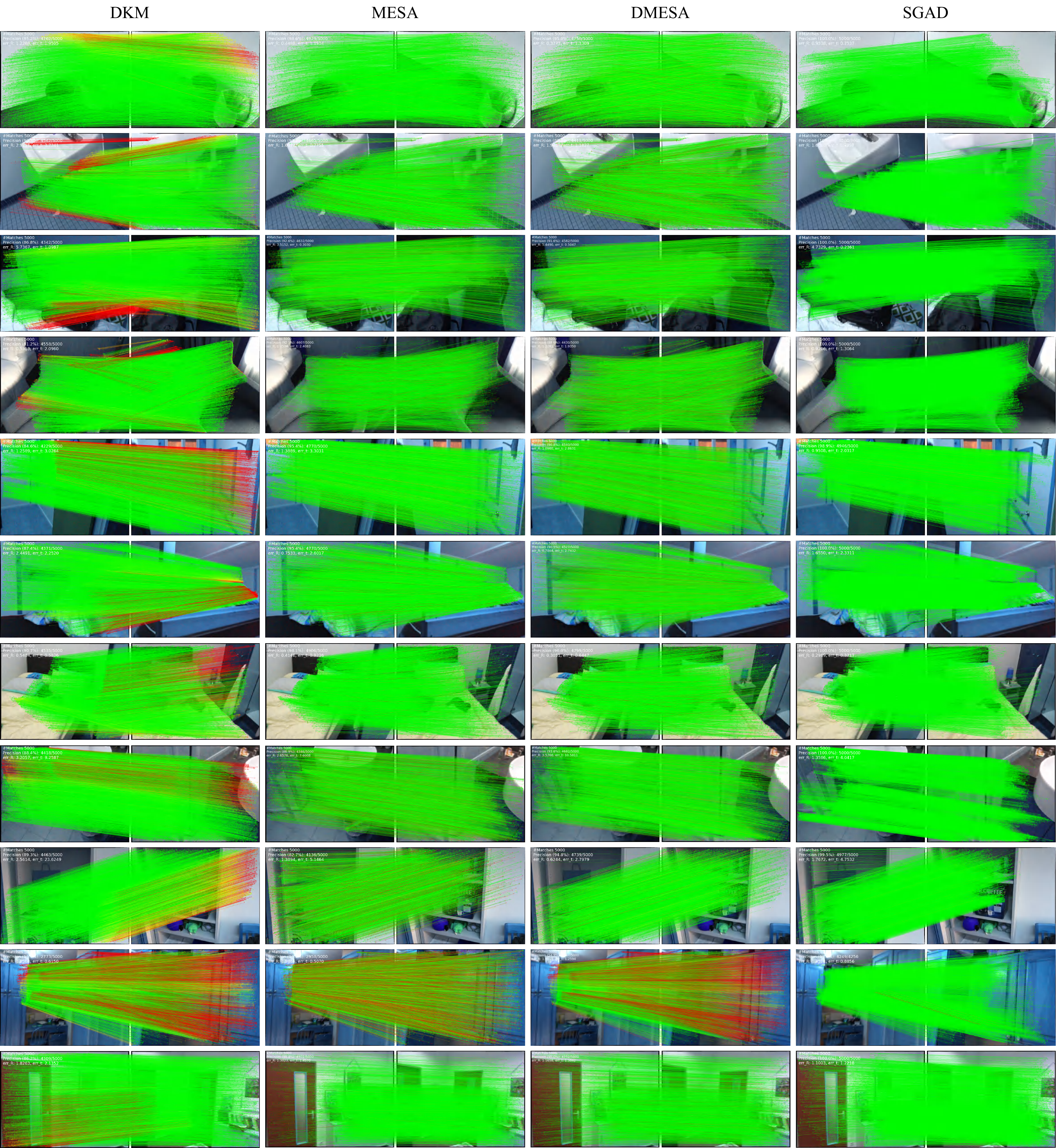}
    \caption{
    Qualitative comparison of matching methods on the ScanNet dataset. 
    Our method (SGAD) is compared against MESA~\cite{mesa}, DMESA~\cite{dmesa}, and DKM~\cite{dkm}.
    To ensure a fair comparison of the upstream area matchers, SGAD, MESA, and DMESA all use DKM as the downstream point matcher, while DKM is also evaluated as a standalone baseline.
    Matches with an epipolar error greater than \(5 \times 10^{-4}\) are highlighted in red. 
    The results show SGAD leads to more correct final matches and lower pose errors.
    }
    \label{fig:suppl_sn}
\end{figure*}

\begin{figure*}[htbp]
    \centering
    \includegraphics[width=\textwidth]{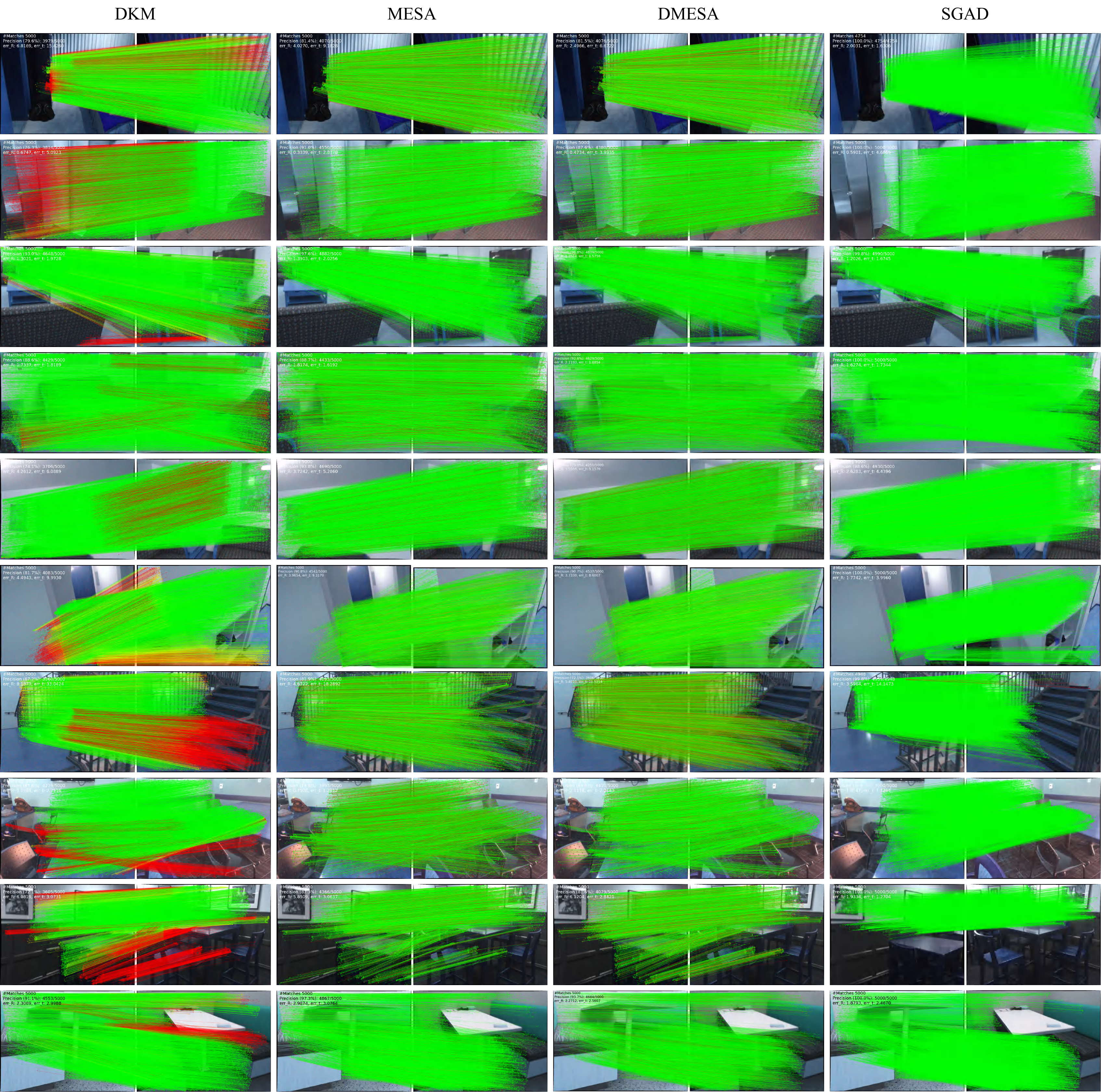}
    \caption{
    Qualitative comparison of matching methods on the ScanNet dataset (continued).
    }
    \label{fig:suppl_sn_continued}
\end{figure*}

Qualitative results for relative pose estimation are presented for the MegaDepth (~\cref{fig:suppl_md,fig:suppl_md_continued}) and ScanNet (~\cref{fig:suppl_sn,fig:suppl_sn_continued}) datasets. 
Following the protocol of LoFTR~\cite{loftr}, we report rotation and translation errors. 
Match precision is visualized by epipolar error, where red indicates errors exceeding the threshold ($1 \times 10^{-4}$ for MegaDepth and $5 \times 10^{-4}$ for ScanNet). 
Across both datasets, our method consistently achieves more correct matches and lower pose errors, highlighting its robustness and accuracy under diverse conditions.

\end{document}